\documentclass[11pt]{article}
\usepackage[final]{acl}
\usepackage{times}
\usepackage{latexsym}
\usepackage{microtype}
\usepackage{inconsolata}
\usepackage{amsmath}
\usepackage{amssymb}
\usepackage{booktabs}
\usepackage{graphicx}
\usepackage{multirow}
\usepackage{array}
\usepackage{xcolor}
\usepackage{url}
\usepackage{hyperref}
\usepackage{algorithm}
\usepackage{algpseudocode}
\usepackage{enumitem}
\usepackage{makecell}
\usepackage{colortbl}
\usepackage{float}
\usepackage{tabularx}
\usepackage[most]{tcolorbox}
\usepackage{listings}

\definecolor{darkblue}{RGB}{29,78,216}
\definecolor{darkgreen}{RGB}{21,128,61}
\definecolor{darkred}{RGB}{185,28,28}

\newcommand{\staropd}{\textsc{STAR-OPD}}
\newcommand{\eabsa}{E-ABSA20K}
\newcommand{\pp}{\texttt{PRODUCT\_PART}}
\newcommand{\pr}{\texttt{PRODUCT}}

\title{STAR-OPD: Structured Aspect-Cascade-Aware
On-Policy Reward Distillation
for ABSA Quadruple Extraction}

\author{
Tong Sun\textsuperscript{1},
Mingyang Ma\textsuperscript{1},
Jiayang Yu\textsuperscript{1} \\
\textsuperscript{1}Alibaba International Digital Commerce Group \\
{\footnotesize \texttt{st422019@alibaba-inc.com}}\\
{\footnotesize \texttt{mingyang.mmy@lazada.com}} \\
{\footnotesize \texttt{yujiayang.jy@alibaba-inc.com}} 
}

\begin{document}
\maketitle

\begin{abstract}
Aspect-based sentiment analysis (ABSA) quadruple extraction requires jointly predicting target, aspect, opinion, and sentiment over reviews that often contain multiple fine-grained sentiment tuples. While large chain-of-thought (CoT) models perform well on this task, distilling them into smaller deployable models remains difficult. We identify a task-specific failure mode in distilled ABSA extraction: student errors at the target–aspect interface create structurally invalid states, such as broken target–aspect bindings and hallucinated targets, which then corrupt downstream predictions. Conventional off-policy distillation is poorly suited to this setting because it trains only on teacher-generated trajectories and provides little supervision on the student-induced structural states that dominate inference. To address this mismatch, we propose \staropd{} (\textbf{ST}ructured \textbf{A}spect-cascade-aware \textbf{O}n-\textbf{P}olicy \textbf{R}eward \textbf{D}istillation), which builds on generic on-policy distillation and instantiates it for ABSA quadruple extraction with cascade-aware, set-structured rewards. STAR-OPD trains on student rollouts and applies set-structured rewards that directly target binding consistency, target grounding, and fine-grained aspect disambiguation. Experiments on \eabsa{} and SemEval-2014 show that \staropd{} consistently outperforms off-policy and general on-policy baselines, reduces target hallucination, and substantially improves performance on structurally hard cases. With Qwen3-4B, \staropd{} substantially narrows the student--teacher gap while improving inference efficiency, highlighting the importance of on-policy structural correction for distilled ABSA extraction.
\end{abstract}

\section{Introduction}
\label{sec:intro}

Aspect-based sentiment analysis (ABSA) quadruple extraction aims to identify structured sentiment tuples of the form \((target, aspect, opinion, sentiment)\).
Compared with simpler ABSA settings, it is substantially more challenging: each review may contain multiple tuples, targets may refer to products or product-parts, and aspects are selected from fine-grained domain taxonomies.
Recent chain-of-thought (CoT) large language models (LLMs) perform strongly on this task, but their inference cost remains prohibitive for high-throughput e-commerce applications, motivating distillation into smaller deployable models.

\begin{figure}[t]
  \centering
  \includegraphics[width=\columnwidth]{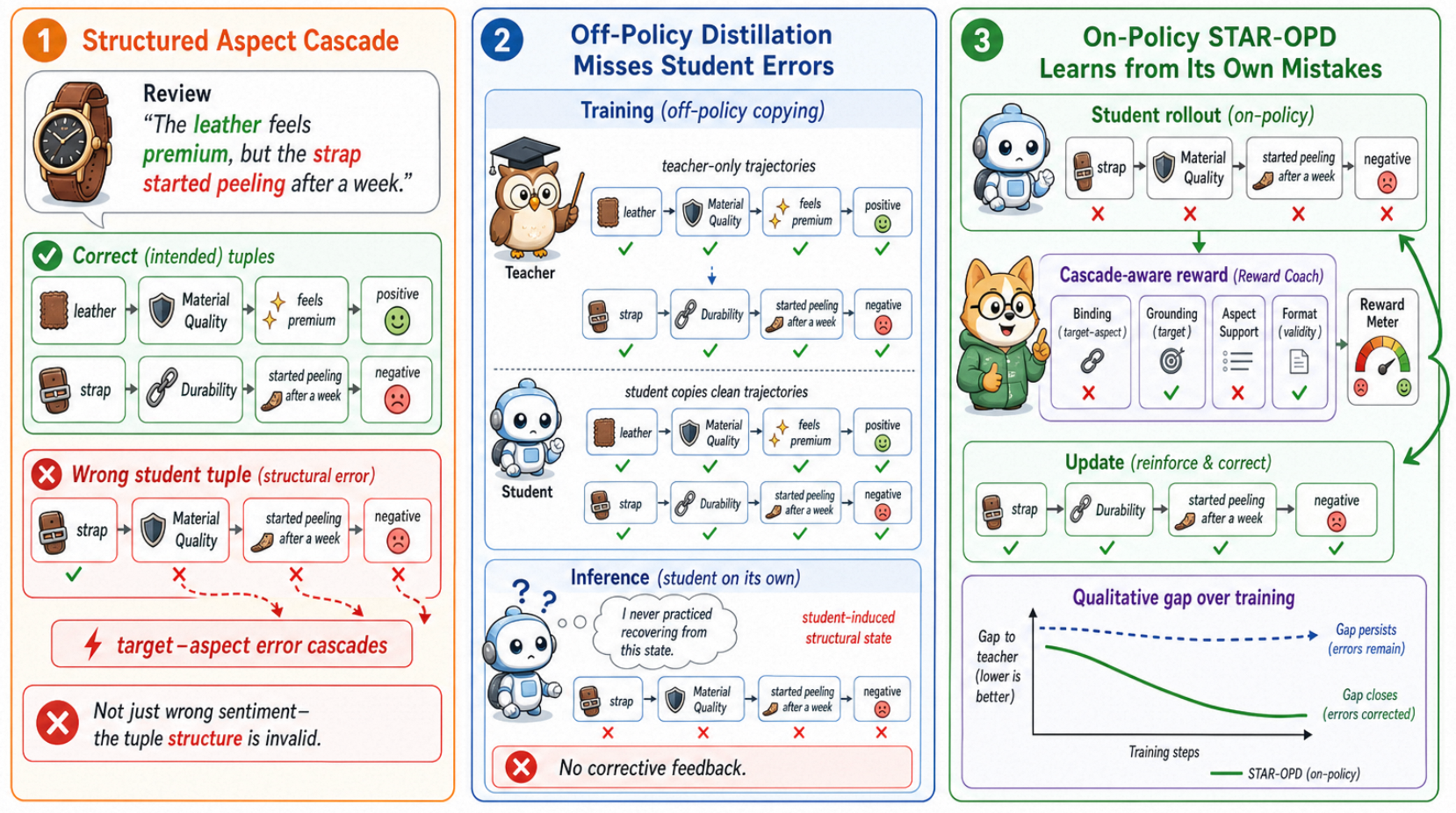}
  \caption{
    Motivation for \staropd{}.
    \textbf{Left:} Errors at the target--aspect interface create structurally invalid tuples and corrupt downstream prediction.
    \textbf{Center:} Off-policy distillation trains only on teacher trajectories and misses student-induced cascade failures.
    \textbf{Right:} On-policy training exposes these states and uses cascade-aware rewards to correct them.
  }
  \label{fig:cascade_motivation}
\end{figure}

We identify a task-specific failure mode in distilled ABSA extraction, which we call \textit{structured aspect cascade}: errors at the target--aspect interface create structurally invalid tuple states, most commonly through broken bindings and hallucinated non-grounded targets.
In a pilot study with a SeqKD distillation baseline, we observe a substantial drop from target-only to target--aspect correctness, suggesting that the main bottleneck lies not in target identification alone, but in preserving valid target--aspect structure.
For example, in ``The leather looks premium, but the strap feels cheap,'' misassigning \textit{strap} to the wrong aspect or hallucinating an unseen product-part can invalidate the entire tuple even if local sentiment words remain plausible.

Conventional \textit{off-policy} distillation is ill-suited to this problem because it supervises the student only on teacher-generated trajectories, which are overwhelmingly structurally valid.
At inference time, however, the student must condition on its own predictions, including incorrect target--aspect bindings and hallucinated targets rarely observed during training.
Because such errors alter tuple structure rather than only local token accuracy, off-policy imitation provides little direct supervision for recovering from the student-induced states that dominate inference-time failure.

To address this mismatch, we propose \staropd{} (\textbf{ST}ructured \textbf{A}spect-cascade-aware \textbf{O}n-\textbf{P}olicy \textbf{R}eward \textbf{D}istillation), which builds on generic on-policy distillation and specializes it for ABSA quadruple extraction.
\staropd{} trains on student rollouts rather than teacher-only trajectories and applies cascade-aware rewards that directly target incorrect target--aspect binding and hallucinated non-grounded targets, with lightweight filtering and sampling as stabilizers.

Experiments on \eabsa{} \citep{2026eabsak} and SemEval-2014 \citep{pontiki-etal-2014-semeval} show that \staropd{} consistently improves over both off-policy baselines and a strong general on-policy baseline.
With Qwen3-4B, it reduces target hallucination from 9.75\% to 7.22\% and yields the largest gains on structurally hard reviews, while also improving deployment efficiency.

\noindent\textbf{Contributions:}
\begin{itemize}[leftmargin=*,noitemsep,topsep=2pt]
  \item We identify \textit{structured aspect cascade}, a task-specific failure mode in distilled ABSA extraction centered on target--aspect binding and target hallucination.
  \item We propose \staropd{}, an instantiation of generic on-policy distillation for ABSA quadruple extraction with cascade-aware rewards for binding consistency, target grounding, and aspect disambiguation.
  \item We show consistent gains over off-policy and generic on-policy baselines on \eabsa{} and SemEval-2014, especially on structurally hard cases.
\end{itemize}

\section{Related Work}
\label{sec:related}

\paragraph{Aspect-Based Sentiment Analysis.}
ABSA has progressed from aspect-level classification \citep{pontiki-etal-2014-semeval}
to triplet \citep{peng2020knowing}
and quadruple extraction \citep{zhang2021aspect,cai2021aspect}.
Recent generative and LLM-based methods \citep{scaria-etal-2024-instructabsa, wang2023chatgpt, hu-etal-2022-improving-aspect}
perform well, but existing quadruple extraction work does not study the structural dependency and distillation-specific error propagation issues that arise in multi-quadruple reviews.

\paragraph{LLM Knowledge Distillation.}
SeqKD \citep{kim2016sequence} distills teacher outputs off-policy and suffers from train--test mismatch.
Recent on-policy methods address this by optimizing on student-generated sequences, including MiniLLM \citep{gu2024minillm}, GKD \citep{agarwal2024gkd}, and DistiLLM \citep{ko2024distillm}.
Subsequent work further extends this framework, including G-OPD \citep{yang2025gopd}, entropy-aware distillation \citep{jin2026entropyaware}, and RLKD \citep{xu2026rlkd}; CoT distillation \citep{magister2022teaching,ho2022large} transfers reasoning traces but has not been studied for structured extraction.
However, these methods are designed for general text generation and optimize generic sequence-level objectives, without modeling field-level binding constraints or dependency structure in quadruple extraction.
Rather than proposing a new generic on-policy distillation principle, our method instantiates this family for ABSA quadruple extraction with set-structured, cascade-aware rewards that provide task-specific structural credit assignment for binding consistency, target grounding, and fine-grained aspect disambiguation.

\paragraph{Error Propagation in Structured Prediction.}
Prior work on exposure bias \citep{bengio2015scheduled}
and error propagation in structured prediction \citep{finkel2006solving}
has shown that training--inference mismatch can amplify upstream mistakes.
We extend this perspective to quadruple extraction distillation, where errors at the target--aspect interface act as a structural bottleneck that corrupts downstream fields within a single quadruple.

\section{Problem Analysis}
\label{sec:analysis}

\subsection{Task Structure and Failure Mode}
\label{sec:gap}

Given a review text $\mathbf{x}$, the goal of ABSA quadruple extraction is to predict a set of sentiment quadruples
\(
\mathcal{Q}(\mathbf{x})=\{(t_i,a_i,o_i,s_i)\}_{i=1}^{K}
\),
where $t_i$ is either a generic product tag or a target mention grounded in the review text, $a_i \in \mathcal{C}$ is a fine-grained aspect category from a closed taxonomy, $o_i$ is an opinion span, and $s_i \in \{\text{pos},\text{neu},\text{neg}\}$ is sentiment polarity.
Quadruple extraction is structurally challenging because each review may contain multiple tuples and the fields are semantically interdependent.
Among these fields, the target--aspect interface is especially critical because it determines how downstream opinion and sentiment should be interpreted within each tuple.
Detailed label definitions, target hierarchy, and dataset statistics are provided in Appendix~\ref{app:data}.

To understand where distilled models fail on this task, we conduct a pilot diagnostic comparison between our strong CoT teacher and a smaller direct model on \eabsa{}-Hard.\footnote{The teacher here is the same high-quality 32B model later used for pseudo-label generation and distillation; preparation details are given in Appendix~\ref{app:teacher}. The smaller direct model is a 4B direct baseline trained without on-policy distillation.}
We find that the smaller model is already close to the teacher when evaluation considers target identification alone (91.0\% vs.\ 97.6\%), a gap of only 6.6 percentage points.
However, once correctness requires preserving each target together with its aspect, the gap widens sharply to 30.5 points (50.6\% vs.\ 81.1\%).
The same pattern continues downstream: when sentiment must also be correct, the gap further increases to 36.1 points (42.5\% vs.\ 78.6\%), and full quadruple correctness remains 36.3 points lower for the smaller model (40.0\% vs.\ 76.3\%).
Viewed relatively, the smaller model retains 93.2\% of teacher performance for target identification alone, but only 62.4\% once correct target--aspect binding is required.
This indicates that the main bottleneck is not target grounding itself, but maintaining valid target--aspect structure under fine-grained aspect ambiguity.

The smaller model also exhibits a noticeably higher target hallucination rate, showing that the problem is not merely choosing the wrong aspect for an existing target, but also entering structurally invalid states with fabricated non-grounded targets.
We refer to this failure pattern as \textit{structured aspect cascade}.
It is centered on the target--aspect interface and manifests primarily as broken target--aspect bindings and hallucinated targets.
Once such an invalid state is entered, downstream fields are conditioned on an incorrect structural interpretation of the tuple, which reduces full-quadruple correctness even when local sentiment expressions remain plausible.

A second pilot with a SeqKD baseline exhibits the same pattern: target identification remains relatively strong, but performance drops sharply once target--aspect correctness is required, and 8.1\% of correctly identified target--aspect pairs still receive an incorrect sentiment label.

Together, these observations suggest that the dominant failure mode of distilled ABSA extraction is not generic sequence noise, but structural corruption at the target--aspect interface.
All values above are micro-averaged F1 unless otherwise specified; detailed statistics and metric definitions are given in Appendix~\ref{app:eval}.

\begin{figure*}[t]
  \centering
  \includegraphics[width=2.0\columnwidth]{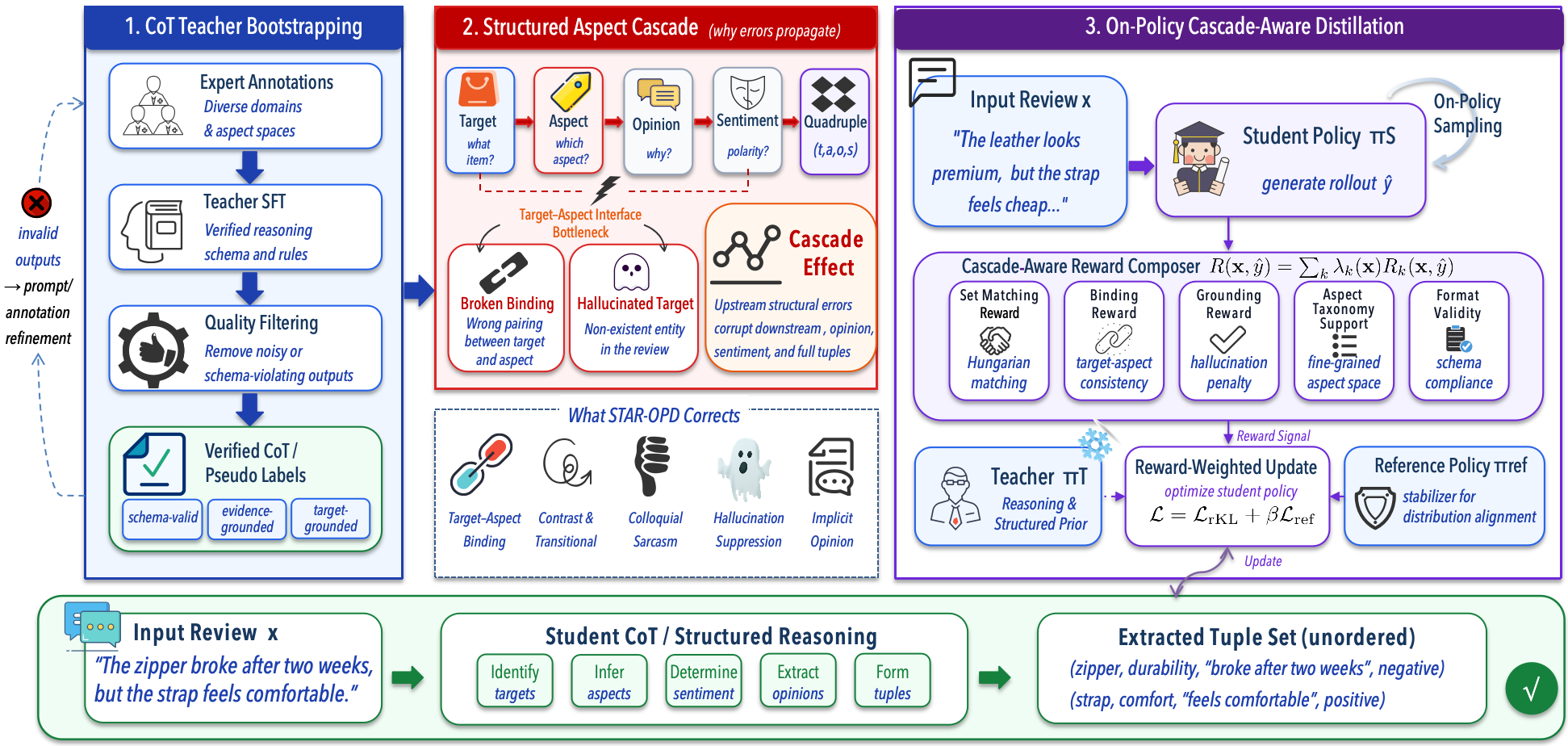}
  \caption{
    Overview of \staropd{}.
    We first prepare a high-quality CoT teacher and generate filtered pseudo-labels for quadruple extraction.
    We then train the student on-policy under its own rollouts, using cascade-aware rewards to correct target--aspect binding errors and hallucinated targets.
  }
  \label{fig:framework}
\end{figure*}

\subsection{Why Off-Policy Distillation Is Insufficient}
\label{sec:offpolicy}

Off-policy distillation supervises the student on teacher-generated trajectories, which are overwhelmingly structurally valid.
At inference time, however, the student conditions on its own predictions and may enter states with incorrect target--aspect bindings or hallucinated targets.
Because these student-induced structural states are largely absent during training, off-policy imitation provides little direct signal for correcting them.
This mismatch is especially harmful in ABSA quadruple extraction, where an early structural mistake changes the semantic interpretation of the remaining fields rather than merely degrading local token accuracy.
An effective solution therefore needs both to expose the student to its own trajectories and to provide feedback at the level of tuple structure, especially around the target--aspect interface.

\section{Method}
\label{sec:method}

\subsection{Overview}

To address the off-policy mismatch identified in Section~\ref{sec:offpolicy}, we propose \staropd{}, which builds on generic on-policy distillation by training on student rollouts and applying structured aspect-cascade-aware rewards at the tuple level, especially for incorrect target--aspect bindings and hallucinated targets.

In practice, we first prepare a high-quality CoT teacher for pseudo-label generation and filtering, then optimize the student on-policy using student rollouts, reverse-KL distillation, and structure-aware rewards, with difficulty-aware sampling and replay as lightweight stabilizers (Appendix~\ref{app:teacher}, \ref{app:train}).

\subsection{On-Policy Cascade-Aware Distillation}

Let $\pi_T$ denote the teacher policy, $\pi_S$ the student policy, and $\pi_{\mathrm{ref}}$ the frozen reference policy initialized from the student before on-policy updates.
Given an input review $\mathbf{x}$, standard off-policy distillation optimizes the student only on teacher-generated trajectories.
In \staropd{}, following generic on-policy distillation, the student is optimized on its own sampled outputs:
\[
\hat{y} \sim \pi_S(\cdot \mid \mathbf{x}).
\]
This exposes training to the same student-induced and often structurally invalid states that occur at inference time, including incorrect target--aspect bindings and hallucinated targets.

We optimize the student with a reward-weighted on-policy distillation objective:
\begin{equation}
  \mathcal{L}
  =
  \mathcal{L}_{\mathrm{rKL}}
  +
  \beta \mathcal{L}_{\mathrm{ref}},
\label{eq:loss}
\end{equation}

where $\alpha(\mathbf{x},\hat{y}) = w(R(\mathbf{x},\hat{y}))$ denotes the reward-derived rollout weight,

\begin{equation}
  \mathcal{L}_{\mathrm{rKL}} =
  \mathbb{E}_{\hat{y} \sim \pi_S}
  \left[
    \alpha(\mathbf{x},\hat{y})
    \log \frac{\pi_S(\hat{y}\mid\mathbf{x})}
    {\pi_T(\hat{y}\mid\mathbf{x})}
  \right],
  \label{eq:rkl}
\end{equation}

and

\begin{equation}
  \mathcal{L}_{\mathrm{ref}} =
  \mathbb{E}_{\hat{y} \sim \pi_S}
  \left[
    \log \frac{\pi_S(\hat{y}\mid\mathbf{x})}
    {\pi_{\mathrm{ref}}(\hat{y}\mid\mathbf{x})}
  \right].
  \label{eq:lref}
\end{equation}

Here $R(\mathbf{x}, \hat{y})$ is a cascade-aware structural reward, and $w(\cdot)$ maps the batch-normalized reward to a non-negative rollout weight (the exact transformation is given in Appendix~\ref{app:reward_weight}), and $\beta$ controls the strength of reference regularization.
Intuitively, higher-reward rollouts receive stronger teacher-aligned updates, while low-reward rollouts contribute less.
The reverse-KL term favors structurally valid teacher-supported outputs, and the reference regularizer stabilizes on-policy updates.

On-policy exposure alone, however, does not specify which structural property of a rollout should be corrected.
We therefore define a cascade-aware reward over student outputs:

\begin{equation}
  R(\mathbf{x}, \hat{y}) = \sum_k \lambda_k(\mathbf{x}) R_k(\mathbf{x}, \hat{y}),
  \label{eq:reward}
\end{equation}

where each component reward targets a distinct structural failure mode of quadruple extraction.
In practice, we normalize the raw reward within each batch and transform it into a non-negative rollout weight.
This design turns on-policy exposure into targeted structural correction: the student is trained on its own trajectories, but updates are biased toward rollouts that better preserve target--aspect consistency and target grounding.

\subsection{Cascade-Aware Rewards}
\label{sec:reward}

Our reward design follows the structural failures identified in Section~\ref{sec:analysis}.
Because ABSA quadruple extraction is evaluated as an unordered set prediction problem, rewards should reflect tuple-set structure rather than generation order.
We therefore compute all reward components after optimal bipartite matching between predicted and gold quadruples, making reward computation invariant to generation order and better aligned with set-level evaluation.
The reward focuses on three dominant issues: incorrect target--aspect binding, hallucinated targets, and ambiguity among fine-grained aspect categories.

\paragraph{Base matching reward.}
We first define a set-level alignment reward that measures overall structural compatibility between predicted and gold quadruples.
Let $\mathcal{Q}^* = \{q_i^*\}_{i=1}^{K^*}$ denote the gold quadruples and
$\hat{\mathcal{Q}} = \{\hat{q}_j\}_{j=1}^{\hat{K}}$ the predicted quadruples.
For each gold--prediction pair, we define
\begin{equation}
\scalebox{0.85}{ 
$S_{ij} =
\frac{1}{3}\Big[
\mathbf{1}[t_i^* = \hat{t}_j]
+
\mathbf{1}[a_i^* = \hat{a}_j]
+
\mathbf{1}[s_i^* = \hat{s}_j]
\Big],$
}
\label{eq:sim}
\end{equation}
and obtain an optimal one-to-one alignment $\mathcal{M}$ using the Hungarian algorithm.
The resulting reward is
\begin{equation}
R_{\text{base}}(\mathbf{x},\hat{y}) =
\frac{1}{K^*}
\sum_{(i,j)\in\mathcal{M}} S_{ij}.
\label{eq:rbase}
\end{equation}
This term provides a global set-level signal for overall tuple quality.

\paragraph{Binding reward.}
Our core reward directly targets the bottleneck at the target--aspect interface:
\begin{equation}
\scalebox{0.8}{ 
$R_{\text{bind}}(\mathbf{x},\hat{y})=
\frac{1}{|\mathcal{M}|+\varepsilon}
\sum_{(i,j)\in\mathcal{M}}
\mathbf{1}[t_i^* = \hat{t}_j]\,
\mathbf{1}[a_i^* = \hat{a}_j].$
}
\label{eq:rpart}
\end{equation}
Unlike $R_{\text{base}}$, $R_{\text{bind}}$ gives credit only when target and aspect are jointly correct, directly targeting the main structural bottleneck under student rollouts.

\paragraph{Hallucination penalty.}
To suppress fabricated non-\pr{} targets, we define a grounded hallucination reward:
\begin{equation}
\scalebox{0.8}{ 
$R_{\text{hall}}(\mathbf{x},\hat{y}) = \mathbb{1}[|\hat{\mathcal{Q}}| > 0] \cdot \left( 1 - \frac{\sum_{\hat{q}_j \in \hat{\mathcal{Q}}} \mathrm{Hall}_t(\hat{q}_j, \mathbf{x})}{|\hat{\mathcal{Q}}|} \right),$
}
\label{eq:rhall}
\end{equation}

where $\mathbb{1}[\cdot]$ is the indicator function, and $\mathrm{Hall}_t(\hat{q}_j, \mathbf{x}) \in \{0, 1\}$ indicates that the predicted quadruple $\hat{q}_j$ contains a non-\pr{} target whose entity span cannot be grounded (i.e., is absent) in the review text $\mathbf{x}$. 

Crucially, the indicator $\mathbb{1}[|\hat{\mathcal{Q}}| > 0]$ prevents the student model from exploiting a trivial reward-hacking shortcut---specifically, generating empty predictions to mathematically bypass the hallucination penalty. This formulation ensures that $R_{\text{hall}}$ acts as a rigorous grounding constraint, assigning a reward of $1.0$ only when the student generates a non-empty set of predictions that are entirely grounded in the source text.

\paragraph{Category reward.}
Hard matching alone provides limited guidance when several aspect labels are semantically adjacent.
To preserve fine-grained teacher supervision within the aspect taxonomy, we add a teacher-support reward over the student-predicted aspect category:
\begin{equation}
\scalebox{0.9}{ 
$R_{\text{cat}}(\mathbf{x},\hat{y}) =
\frac{1}{|\mathcal{M}|+\varepsilon}
\sum_{(i,j)\in\mathcal{M}}
P_T(\hat{a}_j \mid \mathbf{x}, \mathrm{ctx}_j),$
}
\label{eq:rcat}
\end{equation}
where $\hat{a}_j$ is the student-predicted aspect in the aligned tuple, $\mathrm{ctx}_j$ is the student decoding context at which the aspect for the $j$-th predicted tuple is produced, and $P_T(\cdot)$ is the teacher distribution over the aspect taxonomy. In practice, $P_T(\cdot \mid \mathbf{x}, \mathrm{ctx}_j)$ is obtained by a lightweight teacher scoring step under the student rollout context at the aspect prediction step, after restricting and normalizing the logits over the domain taxonomy; details are given in Appendix~\ref{app:train}.
This term complements the discrete binding reward with a softer signal over nearby aspect categories.

\paragraph{Additional stabilizers.}
We further include a lightweight format-validity reward and use input-adaptive weighting, with stronger emphasis on binding consistency and hallucination suppression for reviews identified as likely product-part cases based on teacher pseudo-labels.
The full cascade-aware reward is therefore
\begin{equation}
\scalebox{0.8}{ 
$R(\mathbf{x}, \hat{y}) = \sum_{k \in \{\text{base, bind, hall, cat, fmt}\}} \lambda_k(\mathbf{x}) R_k(\mathbf{x}, \hat{y}).$
}
\label{eq:reward_full}
\end{equation}

where the input-dependent coefficients $\lambda_k(\mathbf{x})$ control the relative contribution of each component.

Among these rewards, $R_{\text{bind}}$ and $R_{\text{hall}}$ are the most important.
The former directly targets the target--aspect interface where structured aspect cascade is centered, while the latter suppresses hallucinated targets that invalidate the entire tuple.
On-policy exposure and cascade-aware rewards are complementary: the former lets the student visit its own erroneous structural states, while the latter provides explicit credit assignment over which structural property of the rollout should be corrected.
Detailed reward weights and training settings are given in Appendix~\ref{app:train}.

Because the reward is rule-based, a natural concern is shortcut behavior; we discuss this further in Appendix~\ref{app:reward_dynamics}.


\section{Experiments}
\label{sec:experiments}

We design our experiments to answer three questions:
(Q1) Does \staropd{} improve overall distilled ABSA extraction quality over off-policy and generic on-policy baselines?
(Q2) Does it specifically reduce the structural failure modes identified in Section~\ref{sec:analysis}, namely the target--aspect bottleneck and target hallucination?
(Q3) Are the gains largest on reviews where structural ambiguity is strongest?

\begin{table*}[t]
  \centering
  \small
  \caption{
    Main results on \eabsa{} and SemEval-2014 (Micro F1).
    Best student in \textbf{bold}; runner-up \underline{underlined}.
    $\dagger$: significantly better than MiniLLM ($p{<}0.05$, paired bootstrap).
  }
  \label{tab:main}
  \setlength{\tabcolsep}{6pt}
  \begin{tabular}{l cccc cc}
    \toprule
    & \multicolumn{4}{c}{\textbf{\eabsa{}}}
    & \multicolumn{2}{c}{\textbf{SemEval-2014}} \\
    \cmidrule(lr){2-5}\cmidrule(lr){6-7}
    \textbf{Method}
    & WomenBags & Dresses & Makeup & Furniture
    & Restaurant & Laptop \\
    \midrule
    \textbf{Teacher}
      & 0.756 & 0.762 & 0.757 & 0.812 & 0.746 & 0.516 \\
    \midrule
    \rowcolor{gray!10}
    \multicolumn{7}{c}{\textbf{Student: Qwen3-4B}} \\
    \midrule
    Direct-SFT
      & 0.685 & 0.691 & 0.692 & 0.720 & 0.685 & 0.470 \\
    SeqKD
      & 0.687 & 0.695 & 0.699 & 0.724 & 0.715 & 0.468 \\
    GKD
      & 0.689 & 0.699 & 0.704 & 0.727 & 0.719 & 0.471 \\
    MiniLLM
      & \underline{0.705} & \underline{0.711}
      & \underline{0.705} & \underline{0.738}
      & \underline{0.728} & \underline{0.481} \\
    G-OPD
      & 0.711 & 0.713 & 0.711 & 0.741 & 0.728 & 0.480 \\
    \textbf{\staropd{}}
      & \textbf{0.739}$^\dagger$ & \textbf{0.750}$^\dagger$
      & \textbf{0.726}$^\dagger$   & \textbf{0.771}$^\dagger$
      & \textbf{0.732}   & \textbf{0.487} \\
    \midrule
    \rowcolor{gray!10}
    \multicolumn{7}{c}{\textbf{Student: Qwen3-1.7B}} \\
    \midrule
    Direct-SFT
       &0.657	 &0.661	 &0.637	 &0.685 & 0.553 & 0.402 \\
    SeqKD
      &0.666	 &0.664	 &0.647	 &0.702 & 0.570 & 0.405 \\
    GKD
      & 0.671 & 0.673 & 0.652 & 0.705 & 0.580 & 0.412 \\
    MiniLLM
      &\underline{0.694}	 &\underline{0.695}	 &\underline{0.660}	 &\underline{0.716}
      & \underline{0.608} & \underline{0.421} \\
    G-OPD
      & 0.699 & 0.702 & 0.662 & 0.717 & 0.607 & 0.421 \\
    \textbf{\staropd{}}
      & \textbf{0.702}$^\dagger$ & \textbf{0.707}$^\dagger$
      & \textbf{0.673}$^\dagger$ & \textbf{0.721}$^\dagger$
      & \textbf{0.614}$^\dagger$ & \textbf{0.425} \\
    \bottomrule
  \end{tabular}
\end{table*}

\subsection{Setup}
\label{sec:setup}

\paragraph{Datasets.}
We evaluate on \eabsa{} \citep{2026eabsak}, a 20K-review benchmark spanning four e-commerce domains, and SemEval-2014 \citep{pontiki-etal-2014-semeval} (Restaurant and Laptop) for cross-domain validation.
We adopt \eabsa{} as the primary benchmark for this study, as its longer reviews and higher tuple density make target--aspect binding errors and target hallucination substantially more frequent than in earlier ABSA settings.

\paragraph{Baselines.}
Unless otherwise specified, all student methods use Qwen3-4B.
We compare against five baselines.
\textbf{Direct-SFT} fine-tunes the student on filtered teacher-produced pseudo-labels as ordinary supervision targets.
\textbf{SeqKD} \citep{kim2016sequence} is the classical off-policy sequence-distillation baseline, training the student to imitate complete teacher-generated output sequences as hard targets. Direct-SFT treats teacher outputs as ordinary supervised targets, whereas SeqKD is reported as the canonical sequence-level off-policy distillation baseline.

\textbf{GKD} \citep{agarwal2024gkd} is a general distillation baseline that combines off-policy and on-policy learning, training the student on both fixed target sequences and self-generated outputs with teacher feedback.

\textbf{MiniLLM} \citep{gu2024minillm} is a strong general on-policy reverse-KL distillation method without task-specific structural rewards.
We also report the \textbf{Teacher} as reference.

\textbf{G-OPD} \citep{yang2025gopd} is a generalized on-policy distillation framework that extends standard OPD with a reward scaling factor and a flexible reference model, enabling stronger student updates and, in some settings, performance beyond the teacher.

GKD, MiniLLM, and G-OPD all represent generic on-policy distillation baselines without ABSA-specific structural reward design.

\paragraph{Evaluation metrics.}
Our primary metric is \textbf{Quad-F1}, with Hungarian matching over target, aspect, and sentiment, followed by opinion evaluation using a fuzzy character-level criterion (CharF1 with threshold $\tau{=}0.5$) to avoid over-penalizing minor boundary deviations.

We additionally report \textbf{T-F1}, \textbf{TA-F1}, \textbf{TAS-F1}, and \textbf{T-Hall} for structural diagnosis (Figure~\ref{fig:field}).
Formal definitions of all metrics are given in Appendix~\ref{app:eval}.

\paragraph{Implementation details.}
Detailed optimization settings, reward-weighting details, hardware configuration, and runtime statistics are reported in Appendix~\ref{app:train}.

\subsection{Main Results}

Table~\ref{tab:main} reports the overall results on \eabsa{} and SemEval-2014.

\staropd{} consistently improves over the off-policy baselines and also outperforms the generic on-policy baselines, including MiniLLM and G-OPD, on the more structurally demanding \eabsa{} benchmark.

On SemEval-2014, the gains over the generic on-policy baselines are smaller, and statistical significance is less consistent.
We attribute this to the simpler review structure and lower tuple density of SemEval, which leave less room for structured cascade and therefore reduce the advantage of task-specific structural rewards.

For Qwen3-4B, \staropd{} substantially narrows the gap to the 32B teacher while retaining the efficiency advantage of the smaller model, and the same trend transfers to the 1.7B student.

\subsection{Reducing Structured Cascade}

\begin{figure}[t]
  \centering
\includegraphics[width=0.95\columnwidth]{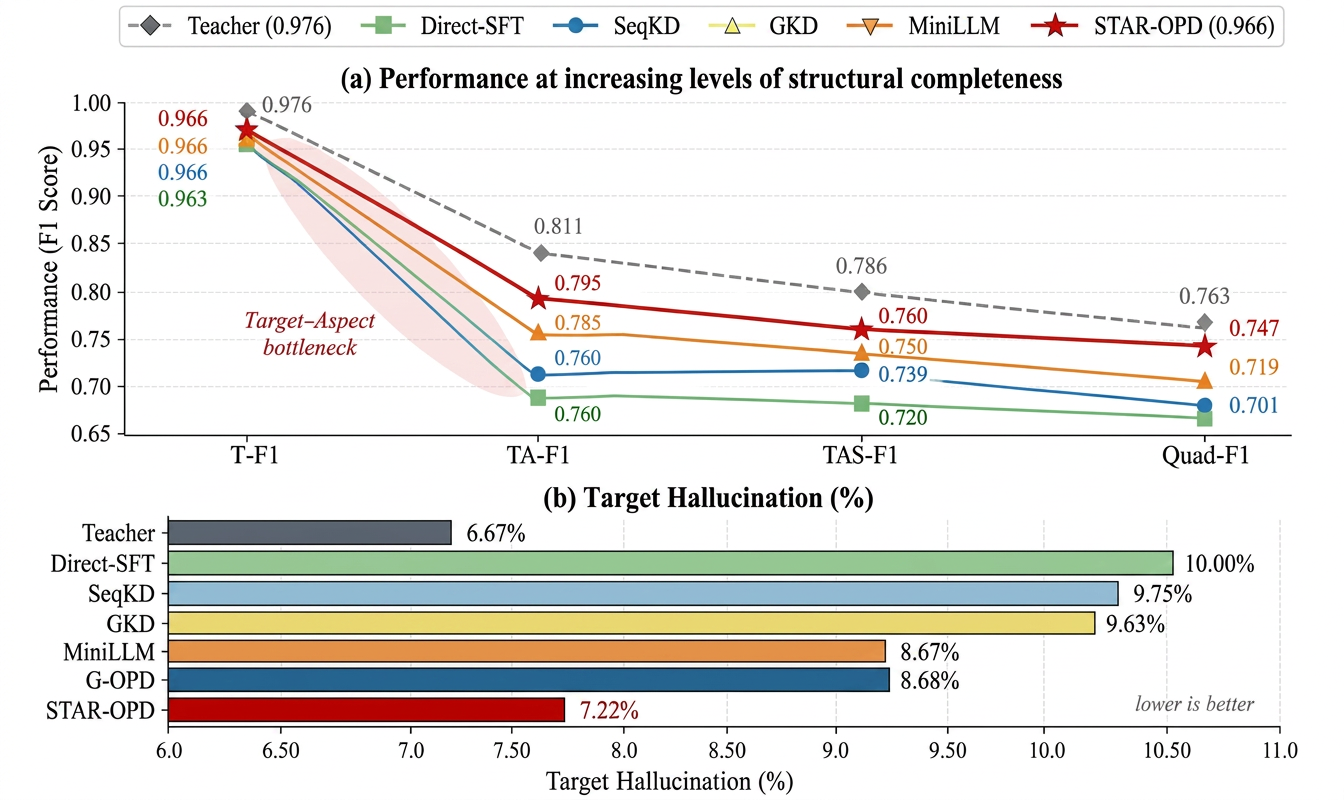}
  \caption{
    Reducing structured aspect cascade on \eabsa{} Test.
    (a) Performance at increasing levels of structural completeness: T-F1, TA-F1, TAS-F1, and Quad-F1.
    (b) Target hallucination rate (T-Hall; lower is better).
    \staropd{} improves all stages of structural correctness, with the largest gain at the target--aspect interface, while reducing hallucinated targets.
  }
  \label{fig:field}
\end{figure}

To test whether \staropd{} reduces the central failure mode identified in Section~\ref{sec:analysis}, Figure~\ref{fig:field} reports field-level performance together with target hallucination rate.

Across all students, target-only performance is already strong, but the much larger drop from T-F1 to TA-F1 confirms that the main bottleneck lies at the target--aspect interface, with further declines to TAS-F1 and Quad-F1 showing downstream effects on sentiment and opinion prediction.

Compared with SeqKD, \staropd{} improves every stage of this progression, with the largest gain at the target--aspect interface, substantially reducing the T--TA gap.

Figure~\ref{fig:field}(b) further shows that \staropd{} lowers target hallucination from 9.75\% to 7.22\%, suggesting improved structural validity rather than merely surface-level imitation.

\subsection{On-Policy and Reward Ablations}
\begin{table}[t]
  \centering
  \small
  \caption{
Ablation of on-policy design choices and cascade-aware rewards on \eabsa{} Test.
F1 denotes micro Quad-F1 over the four \eabsa{} test domains.
$\Delta$F1 is relative to full \textbf{\staropd{}}.
}
  \label{tab:ablation}
  \setlength{\tabcolsep}{4pt}
  \begin{tabular}{lrc}
    \toprule
    \textbf{Variant}
    & \textbf{F1}
    & \textbf{$\Delta$F1} \\
    \midrule
    \textbf{\staropd{}} (full) & 0.747 & ---    \\
    \midrule
    \multicolumn{3}{l}{\textit{On-policy mechanism}} \\
    \;\;w/o On-Policy             & 0.702 & $-$0.045 \\
    \;\;w/o Rewards               & 0.716 & $-$0.031 \\
    \midrule
    \multicolumn{3}{l}{\textit{Reward decomposition}} \\
    \;\;Collapsed scalar reward   & 0.734 & $-$0.013 \\
    \midrule
    \multicolumn{3}{l}{\textit{Cascade-aware rewards}} \\
    \;\;w/o $R_\text{bind}$       & 0.729 & $-$0.018 \\
    \;\;w/o $R_\text{hall}$       & 0.737 & $-$0.010 \\
    \;\;w/o $R_\text{cat}$        & 0.738 & $-$0.009 \\
    \;\;w/o $R_\text{fmt}$        & 0.735 & $-$0.012 \\
    \;\;w/o $R_\text{adap}$       & 0.737 & $-$0.010 \\
    \midrule
    \multicolumn{3}{l}{\textit{Matching strategy}} \\
    \;\;Greedy (vs.\ Hungarian)   & 0.735 & $-$0.012 \\
    \bottomrule
  \end{tabular}
\end{table}

Table~\ref{tab:ablation} evaluates the contributions of on-policy exposure and cascade-aware rewards.

Removing on-policy training causes the largest drop, confirming that student rollouts are essential for covering the structurally invalid states absent from off-policy teacher supervision.
Removing reward shaping and retaining only plain on-policy distillation also leads to a clear degradation, showing that on-policy exposure alone is not sufficient without explicit structural credit assignment.

We further compare against a collapsed scalar reward variant that merges the main structural signals into a single outcome-level score without component-specific weighting (Appendix~\ref{app:collapsed_reward}).
Its weaker performance indicates that decomposing binding, grounding, and category support provides more effective structural credit assignment than a single undifferentiated reward.

Among reward terms, removing the binding reward $R_{\text{bind}}$ causes the largest performance drop, showing that correcting target--aspect binding errors is central to the method.
Removing $R_{\text{hall}}$, $R_{\text{cat}}$, $R_{\text{fmt}}$, or $R_{\text{adap}}$ also degrades performance, indicating that suppressing non-grounded targets, preserving fine-grained aspect supervision, stabilizing format validity, and emphasizing structurally difficult cases are all beneficial.

Finally, replacing Hungarian matching with greedy matching also degrades performance, showing that order-invariant set-level alignment is important for assigning structurally meaningful rewards.

These results show that on-policy exposure and structure-aware rewards are complementary: the former exposes student error states, and the latter provides explicit structural credit assignment.
This is consistent with the gap between MiniLLM and \staropd{} in Table~\ref{tab:main}.

\subsection{Performance on Hard Cases}

We further evaluate on \eabsa{}-Hard, where structural ambiguity most strongly amplifies cascade failures.
Table~\ref{tab:hard} shows that \staropd{} achieves the best student performance on all hard subsets, with the largest gains on product-part cases and consistent improvements on contrastive and implicit-opinion reviews.
Generic on-policy baselines such as GKD and G-OPD also improve over off-policy distillation on these subsets, but remain less effective than \staropd{} on the most structure-sensitive cases.

\begin{table}[t]
  \centering
  \small
  \caption{
    Hard-sample F1 on \eabsa{}-Hard.
    PP: product-part mentions;
    Sarc.: colloquial sarcasm;
    Cont.: contrastive sentences;
    Impl.: implicit opinion expressions.
  }
  \label{tab:hard}
  \setlength{\tabcolsep}{4pt}
  \begin{tabular}{lcccc}
    \toprule
    \textbf{Method}
    & \textbf{PP}
    & \textbf{Sarc.}
    & \textbf{Cont.}
    & \textbf{Impl.} \\
    \midrule
    \rowcolor{gray!10}\textbf{Teacher}
      & 0.687 & 0.717 & 0.722 & 0.723 \\
    \midrule
    Direct-SFT  & 0.632 & 0.664 & 0.657 & 0.679 \\
    SeqKD       & 0.633 & 0.678 & 0.671 & 0.680 \\
    GKD       & 0.635 & 0.678 & 0.672 & 0.683 \\
    MiniLLM     & 0.643 & 0.689 & 0.687 & 0.699 \\
    G-OPD     & 0.644 & 0.692 & 0.686 & 0.709 \\
    \textbf{\staropd{}}
      & \textbf{0.665} & \textbf{0.705}
      & \textbf{0.699} & \textbf{0.710} \\
    \bottomrule
  \end{tabular}
\end{table}

Because aggregate scores do not reveal how different distillation strategies fail, Figure~\ref{fig:qual_cases} presents a representative product-part example:
SeqKD preserves the grounded part target but drifts in target--aspect binding,
MiniLLM generates a plausible yet non-grounded target,
and \staropd{} restores the intended grounded target together with its fine-grained aspect.

\begin{figure}[t]
\centering
\footnotesize
\setlength{\tabcolsep}{3pt}
\renewcommand{\arraystretch}{1.05}
\begin{tabularx}{\columnwidth}{>{\raggedright\arraybackslash}p{0.16\columnwidth} X}
\toprule
\rowcolor{gray!10}
\multicolumn{2}{c}{\textbf{Qualitative case: product-part (womenbags)}} \\
\midrule
\textbf{Review} &
The bag looks elegant, but the zipper got stuck on the second day. \\
\textbf{Gold} &
(\texttt{PRODUCT:bag}, \texttt{Design Aesthetic}, \textit{looks elegant}, \texttt{pos}); 
(\texttt{\pp{}:zipper}, \texttt{Hardware Quality}, \textit{got stuck}, \texttt{neg}) \\
\textbf{SeqKD} &
(\texttt{\pp{}:zipper}, \textcolor{darkred}{\texttt{Ease of Use}}, \textit{got stuck}, \texttt{neg})
{\footnotesize\textcolor{darkred}{ [binding drift]}} \\
\textbf{MiniLLM} &
(\textcolor{darkred}{\texttt{\pp{}:handle}}, \texttt{Hardware Quality}, \textit{got stuck}, \texttt{neg})
{\footnotesize\textcolor{darkred}{ [hallucinated target]}} \\
\textbf{\staropd{}} &
(\texttt{\pp{}:zipper}, \texttt{Hardware Quality}, \textit{got stuck}, \texttt{neg})
{\footnotesize\textcolor{darkgreen}{ [correct grounding + binding]}} \\
\bottomrule
\end{tabularx}
\caption{
Representative qualitative case from \eabsa{}-Hard.
SeqKD preserves the grounded target but drifts in aspect binding, whereas MiniLLM generates a plausible yet non-grounded part target.
\staropd{} restores both correct grounding and part-specific aspect binding.
Additional cases are given in Appendix~\ref{app:qual_cases_app}.
}
\label{fig:qual_cases}
\end{figure}

The qualitative example in Figure~\ref{fig:qual_cases} highlights how the baselines fail differently under product-part ambiguity.
SeqKD preserves the observed part mention but drifts to a neighboring aspect, whereas MiniLLM produces a plausible yet non-grounded part target.
In contrast, \staropd{} recovers the intended grounded target together with its part-specific aspect, illustrating why its gains are largest on hard subsets where target granularity and aspect assignment are tightly coupled.

\section{Conclusion}
\label{sec:conclusion}

We studied why distillation is brittle for ABSA quadruple extraction and found that the main challenge is not teacher imitation alone, but recovery from student-induced structural states.
In this task, the most damaging failures are centered on the target--aspect interface, where broken bindings and hallucinated targets degrade downstream tuple correctness.
\staropd{} addresses this mismatch by building on generic on-policy distillation and adding cascade-aware rewards for binding consistency and target grounding.
Across \eabsa{} and SemEval-2014, this design consistently improves distilled students, reduces hallucination, and yields the largest gains on structurally hard reviews.

More broadly, our results suggest that distillation for structured extraction should be designed around inference-time student states rather than teacher-only trajectories.
Our rewards operate at the structured outcome level rather than as token-level process rewards, which is natural for unordered quadruple extraction.
Exploring finer-grained process-style rewards is an interesting direction for future work.

\section{Limitations}

Our study has several limitations. First, although we evaluate on both \eabsa{} and SemEval-2014, the analysis is centered on ABSA quadruple extraction, especially in e-commerce reviews, so the generality of \staropd{} to other structured extraction settings remains to be validated. Second, our reward design operates at the structured outcome level after generation rather than as a token-level or process-level reward, which may limit earlier intervention during decoding. Third, the method depends on the quality of the teacher and pseudo-label filtering pipeline, and our hard-subset analyses are intended as diagnostic evidence of failure modes rather than the sole basis for broad statistical claims.

\paragraph{Potential Risks.}
We do not identify severe direct risks beyond those typical of sentiment analysis systems, but deployment may amplify annotation biases or residual errors in target--aspect binding and target grounding. In practical settings, such errors could lead to incorrect summaries of user opinions or mischaracterization of product attributes, so the method is better viewed as a research and assistive analysis tool rather than a fully autonomous high-stakes decision system.


\bibliography{main}

\appendix

\section{Task and Dataset Details}
\label{app:data}

\subsection{Full Task Definition}

Given a review text $\mathbf{x}$, the goal of ABSA quadruple extraction is to predict a set of sentiment quadruples
\[
\mathcal{Q}(\mathbf{x}) = \{q_i\}_{i=1}^{K}, \qquad
q_i = (t_i, a_i, o_i, s_i),
\]
where each quadruple consists of a target $t_i$, an aspect $a_i$, an opinion span $o_i$, and a sentiment label $s_i$.

The target field follows a structured hierarchy:
 \begin{equation}
\begin{aligned}
t_i \in \{&
\texttt{PRODUCT},\;
\texttt{PRODUCT:}\textit{xx},\\
&\texttt{\pp{}:}\textit{xx}
\}.
\end{aligned}
\end{equation} 

where the entity \textit{xx} must appear verbatim in the review text.
The aspect field $a_i \in \mathcal{C}$ is selected from a closed domain taxonomy with up to $|\mathcal{C}|{=}42$ fine-grained categories depending on the domain.
The opinion field $o_i$ is a minimal contiguous span in the input, and the sentiment field
$s_i \in \{\text{pos}, \text{neu}, \text{neg}\}$.

Compared with simpler ABSA formulations, quadruple extraction is structurally more challenging because each review may contain multiple quadruples and each quadruple combines several heterogeneous prediction subproblems, including entity identification for targets, taxonomy classification for aspects, span extraction for opinions, and polarity classification for sentiment.
In particular, the target--aspect interface forms the key structural bottleneck linking target grounding to downstream sentiment and opinion interpretation, making quadruple extraction especially vulnerable to structural error propagation.

\subsection{Dataset Statistics}

We conduct experiments on \eabsa{} \citep{2026eabsak}, a 20K-review benchmark spanning four e-commerce domains, and on SemEval-2014 \citep{pontiki-etal-2014-semeval} for cross-domain validation.
The \eabsa{} benchmark is substantially more challenging than earlier ABSA datasets because reviews are longer and contain multiple sentiment quadruples.
On average, each review in \eabsa{} contains 6.0 quadruples, creating substantially more opportunities for target confusion, aspect ambiguity, and structural binding errors than shorter benchmark settings.

We additionally construct \eabsa{}-Hard for fine-grained diagnosis.
This subset contains reviews exhibiting at least one structurally difficult phenomenon, including product-part mentions, sarcasm cases, contrastive sentences, implicit opinion expressions.
These phenomena are particularly useful for analyzing the failure modes of distilled models because they amplify aspect ambiguity, target--aspect binding difficulty, and target hallucination.

\subsection{Construction of \eabsa{}-Hard}

We define \eabsa{}-Hard as the subset of reviews that contain at least one challenging phenomenon known to increase structural ambiguity in quadruple extraction.
The subset includes the following categories:

\begin{itemize}[leftmargin=*,noitemsep,topsep=2pt]
    \item \textbf{Product-part mentions}: reviews containing fine-grained product-part targets, which require the model to preserve target identity at a more specific level than the generic product tag.
    \item \textbf{Colloquial sarcasm}: reviews whose sentiment is expressed indirectly or ironically, often weakening the reliability of surface lexical cues.
    \item \textbf{Contrastive sentences}: reviews containing discourse structures such as \textit{but}, where sentiment orientation may reverse across clauses and must be resolved in an aspect-sensitive way.
    \item \textbf{Implicit opinion expressions}: reviews in which polarity must be inferred without explicit sentiment words, increasing reliance on aspect semantics and context.
\end{itemize}

These categories are used only for diagnostic evaluation and hard-case analysis.
They are not required by the proposed method, but help reveal where structured aspect cascade is most severe.
Because these subsets are smaller and partially overlapping, we treat the corresponding results as diagnostic rather than as the primary basis for significance claims.

\section{Evaluation Details}
\label{app:eval}

\subsection{Field-Level Matching Metrics}

To analyze structural errors at different levels of completeness, we report four cumulative matching metrics:
target-only F1 (T-F1), target--aspect F1 (TA-F1), target--aspect--sentiment F1 (TAS-F1), and quadruple F1 (Quad-F1).

Let $\mathcal{Q}^* = \{q_i^*\}_{i=1}^{K^*}$ denote the gold quadruples and
$\hat{\mathcal{Q}} = \{\hat{q}_j\}_{j=1}^{\hat{K}}$ denote the predicted quadruples, where each quadruple has the form
$q=(t,a,o,s)$.

\paragraph{T-F1.}
A prediction is counted as correct under T-F1 if its target field matches a gold target after one-to-one alignment between predicted and gold quadruples.
This metric measures target identification in isolation.

\paragraph{TA-F1.}
A prediction is counted as correct under TA-F1 if both the target and aspect fields match after alignment.
This metric directly reflects the quality of target--aspect binding and is therefore central to diagnosing the bottleneck at the target--aspect interface.

\paragraph{TAS-F1.}
A prediction is counted as correct under TAS-F1 if the target, aspect, and sentiment fields all match after alignment.
This metric captures structural correctness after downstream sentiment resolution.

\paragraph{Quad-F1.}
A prediction is counted as correct under Quad-F1 if the target, aspect, sentiment, and opinion fields all match after alignment, where opinion matching is evaluated with a fuzzy character-level criterion described below.
This is the primary evaluation metric used in the main experiments.

All F1 scores are reported as micro-averaged set-level F1 across the evaluation corpus.

\subsection{Hungarian Matching for Set Prediction}

Because each review may contain multiple predicted and gold quadruples, evaluation is performed with one-to-one bipartite matching rather than by comparing outputs in generation order.
We compute an optimal alignment between gold and predicted quadruples using the Hungarian algorithm.

For a gold quadruple $q_i^*$ and a predicted quadruple $\hat{q}_j$, we define the structural similarity
\begin{equation}
S_{ij}
=
\frac{1}{3}
\Big(
\mathbf{1}[t_i^*=\hat{t}_j]
+
\mathbf{1}[a_i^*=\hat{a}_j]
+
\mathbf{1}[s_i^*=\hat{s}_j]
\Big),
\end{equation} 
and obtain an optimal one-to-one matching
\(
\mathcal{M}\subseteq \mathcal{Q}^*\times \hat{\mathcal{Q}}
\)
that maximizes total similarity.

After matching, precision, recall, and F1 are computed under the relevant correctness criterion:
target-only matching for T-F1,
joint target--aspect matching for TA-F1,
joint target--aspect--sentiment matching for TAS-F1,
and full quadruple matching for Quad-F1.

We intentionally exclude opinion spans from the matching score used by the Hungarian algorithm.
Because opinion extraction is evaluated with a fuzzy character-level criterion, incorporating it directly into the alignment objective may introduce unstable matches under minor boundary variation.
Using target, aspect, and sentiment for alignment yields a more stable structural correspondence between predicted and gold quadruples, while opinion correctness is still enforced afterward in Quad-F1 through the CharF1 threshold.
Thus, opinion is excluded from alignment but not from final quadruple evaluation.

\subsection{Opinion Matching}

Opinion spans are evaluated with character-level F1 (CharF1), which is more robust than exact span matching to minor boundary deviations.
For a gold opinion span $o_i^*$ and a predicted opinion span $\hat{o}_j$, we compute
\[
\mathrm{CharF1}(o_i^*, \hat{o}_j),
\]
the character-level F1 score between the two spans after normalization.

An aligned prediction is counted as opinion-correct if its opinion CharF1 exceeds a fixed threshold $\tau$.
Accordingly, a matched pair $(q_i^*, \hat{q}_j)$ is counted as correct under Quad-F1 if
\[
t_i^*=\hat{t}_j,\quad
a_i^*=\hat{a}_j,\quad
s_i^*=\hat{s}_j,
\]
and
\[
\mathrm{CharF1}(o_i^*, \hat{o}_j) \ge \tau.
\]

Unless otherwise specified, we use $\tau=0.5$ in all experiments.

\subsection{Target Hallucination Rate (T-Hall)}

Target hallucination rate (T-Hall) measures the proportion of predicted quadruples whose non-\pr{} target entity does not appear in the input review text.
This metric is designed to capture a structurally severe failure mode in distilled quadruple extraction, where the model fabricates a target that cannot be grounded in the source review.

Formally, let $\hat{\mathcal{Q}}=\{\hat{q}_j\}_{j=1}^{\hat{K}}$ be the predicted quadruples for input $\mathbf{x}$, and let $\hat{t}_j$ denote the target field of $\hat{q}_j$.
We define an indicator
\[
\mathrm{Hall}_t(\hat{q}_j,\mathbf{x}) =
\begin{cases}
1, & 
\begin{aligned}[t]
&\hat{t}_j \neq \texttt{PRODUCT} \\
&\land\ \mathrm{ent}(\hat{t}_j) \notin \mathrm{text}(\mathbf{x}),
\end{aligned} \\
0, & \text{otherwise.}
\end{cases}
\]

where $\mathrm{ent}(\hat{t}_j)$ extracts the entity span from the predicted target and
$\mathrm{text}(\mathbf{x})$ denotes the normalized review text.

The target hallucination rate is then
 \begin{equation}
\mathrm{T\mbox{-}Hall}(\mathbf{x})=
\frac{\left|\left\{\hat{q}_j:\mathrm{Hall}_t(\hat{q}_j,\mathbf{x})=1\right\}\right|}
     {|\hat{\mathcal{Q}}|+\varepsilon},
\end{equation} 
where $\varepsilon$ is a small constant to avoid division by zero.
Corpus-level T-Hall is obtained by averaging over the evaluation set.

For targets of the form \texttt{PRODUCT:}\textit{xx} or \texttt{\pp{}:}\textit{xx}, the entity span \textit{xx} is extracted and checked against the review text using exact substring matching after lowercasing and whitespace normalization.
A lower T-Hall indicates that predicted targets are better grounded in the review text.



\section{Additional Analysis of Structural Failure Modes}
\label{app:cascade}

We observe that structured aspect cascade in distilled quadruple extraction is centered on the target--aspect interface and manifests primarily in two ways: broken target--aspect bindings and hallucinated targets.

\paragraph{Broken target--aspect bindings and downstream corruption.}
Because the fields in a quadruple are structurally interdependent, an incorrect aspect assignment changes how the target is interpreted and can degrade downstream sentiment and opinion prediction.
This explains why the largest performance drop occurs at the transition from target-only to target--aspect matching, and why additional degradation remains at the TAS-F1 and Quad-F1 levels even after partial structural correctness is achieved.

\paragraph{Hallucinated targets.}
In addition to misclassifying existing targets, small distilled models may fabricate non-\pr{} entities that do not appear in the review text, especially at the product-part level.
Such hallucinations are structurally severe: once the target itself is invalid, the entire quadruple becomes spurious and no downstream field can be correct.
In the SeqKD baseline, the target hallucination rate reaches 9.75\%, compared with 6.67\% for the teacher (see Appendix~\ref{app:eval} for the formal definition of T-Hall).

Taken together, these observations show that distilled quadruple extraction fails in a distinctly structured way.
The main issue is not generic sequence noise, but student errors that create invalid structural states at the target--aspect interface and thereby corrupt downstream tuple interpretation.

\section{Teacher Preparation and Quality Control}
\label{app:teacher}

\subsection{CoT Teacher Fine-Tuning}

We use a Qwen3-32B model as the teacher.
To improve the reliability of pseudo-label generation, we fine-tune the teacher with chain-of-thought (CoT) supervision so that it explicitly reasons over target identification, aspect assignment, opinion extraction, and sentiment resolution before producing the final quadruples.
Only the final quadruple outputs are used as task targets for distillation; teacher reasoning traces are not distilled.
Teacher distributions may still be queried during training for reward computation (e.g., aspect-category support).

The teacher is trained only on manually annotated training data.
Development annotations are used for model selection, while test annotations are never used for teacher adaptation, pseudo-label generation, reward tuning, or student training.

The resulting model is referred to as \textbf{Teacher} in the main paper.

\subsection{Agreement with Expert Annotations}

To ensure that the teacher provides reliable supervision, we compare teacher predictions against human expert annotations on a held-out subset of \eabsa{}-Hard.
The subset was independently annotated by two domain annotators who were blind to model outputs.
We report teacher--human agreement against the adjudicated labels.
Human--human agreement on the same subset was $\kappa = 0.89$, and disagreements were resolved by discussion with a third annotator.

Table~\ref{tab:teacher_quality} reports agreement at the field level as well as overall Cohen's $\kappa$.
The \textbf{Teacher} achieves substantially higher agreement than a smaller direct model and reaches $\kappa=0.92$, which we treat as sufficient quality for pseudo-label generation.

These results support the use of the CoT teacher as the source of pseudo-label supervision in \staropd{}.

\begin{table}[h]
  \centering
  \small
  \caption{
    Teacher quality against human expert annotations.
    $\kappa$ denotes Cohen's Kappa.
  }
  \label{tab:teacher_quality}
  \setlength{\tabcolsep}{4pt}
  \begin{tabular}{lcc}
    \toprule
    \textbf{Field} & \textbf{32B CoT} & \textbf{4B Direct} \\
    \midrule
    Target agreement    & 97.5\% & 94.2\% \\
    Aspect agreement    & 93.8\% & 81.6\% \\
    Opinion agreement   & 94.1\% & 88.3\% \\
    Sentiment agreement & 95.3\% & 86.7\% \\
    Overall $\kappa$    & 0.92   & 0.81   \\
    \bottomrule
  \end{tabular}
\end{table}

\subsection{Pseudo-Label Generation and Filtering}

Using the validated teacher, we generate pseudo-labels for the student training split.
Because teacher outputs may still contain occasional formatting errors or structurally invalid extractions, we apply lightweight filtering before distillation.

We retain only instances satisfying the following constraints:
\begin{itemize}[leftmargin=*,noitemsep,topsep=2pt]
    \item \textbf{Well-formed quadruples:} each prediction must follow the required output schema and contain all four fields.
    \item \textbf{Valid aspect labels:} the predicted aspect must belong to the domain-specific taxonomy $\mathcal{C}$.
    \item \textbf{Valid sentiment labels:} the predicted sentiment must be one of \{\texttt{pos}, \texttt{neu}, \texttt{neg}\}.
    \item \textbf{Grounded non-\pr{} targets:} when the predicted target is not \texttt{PRODUCT}, the corresponding entity span must appear in the review text.
    \item \textbf{Duplicate removal:} repeated identical quadruples for the same review are collapsed into a single instance.
\end{itemize}

This filtering step is intended only to remove structurally invalid supervision.
The correction of student-induced structural errors is still handled during on-policy training with cascade-aware rewards.
After filtering, we retain 70\% of teacher-generated training instances overall.
The domain-wise retention rates are broadly similar, remaining within 66\%--73\% across the four domains rather than concentrating on a single domain or review type.
Pseudo-labels are generated only for the student training split.
No evaluation annotations from the validation or test sets are used in pseudo-label generation or filtering.
The \eabsa{}-Hard subsets are used only for diagnostic evaluation and never for teacher or student training.

\section{Training Details}
\label{app:train}

\subsection{Reward Weights and Input-Adaptive Weighting}

The total reward is defined as
\[
R(\mathbf{x}, \hat{y}) = \sum_k \lambda_k(\mathbf{x}) R_k(\mathbf{x}, \hat{y}),
\]
where the component rewards include base matching, target--aspect binding, hallucination suppression, category supervision, and format validity.

In practice, we assign larger relative weight to the binding reward and the hallucination penalty, since these two terms directly target the dominant structural failures identified in Section~\ref{sec:analysis}.
The category reward is given moderate weight to preserve fine-grained aspect supervision, while the format reward receives only a small weight as a stabilizing term.
All reward weights are selected on a held-out validation set.

\begin{table}[t]
\centering
\small
\caption{Reward weights used in STAR-OPD. Default weights are used unless the input is identified as a likely \pp{} case based on teacher pseudo-labels.}
\label{tab:reward_weights}
\begin{tabular}{lcc}
\toprule
\textbf{Reward term} & \textbf{Default} & \textbf{\pp{} case} \\
\midrule
$\lambda_{\text{base}}$ & 0.20 & 0.15 \\
$\lambda_{\text{bind}}$ & 0.30 & 0.35 \\
$\lambda_{\text{hall}}$ & 0.25 & 0.30 \\
$\lambda_{\text{cat}}$  & 0.20 & 0.15 \\
$\lambda_{\text{fmt}}$  & 0.05 & 0.05 \\
\bottomrule
\end{tabular}
\end{table}

We selected these weights on the validation set by first fixing a stable default configuration and then varying each coefficient within a narrow range while monitoring Quad-F1, TA-F1, and T-Hall.
We found the method to be relatively stable under moderate ($\pm 0.05$) perturbations around the final values.
We use the same default reward weights across all domains rather than tuning them separately per domain.

To avoid over-tuning a five-way reward decomposition, we used a constrained local grid around a stable default configuration rather than an exhaustive combinatorial search.
Specifically, we first fixed a default profile consistent with the structural diagnosis in Section~\ref{sec:analysis}, then varied one coefficient at a time while monitoring Quad-F1, TA-F1, and T-Hall on the validation set.
This narrow search strategy was chosen to reduce the risk of overfitting to a particular validation distribution.
In practice, we did not observe severe instability from interactions among reward terms; the main trade-off was between stronger hard-constraint rewards ($R_{\text{bind}}, R_{\text{hall}}$) and softer category supervision ($R_{\text{cat}}$).
We intentionally kept the final weights simple and low-dimensional so that the method remains interpretable and easy to reproduce.

We further use lightweight input-adaptive weighting.
For reviews identified as likely product-part cases based on teacher pseudo-labels, we increase the relative emphasis on binding consistency and hallucination suppression, since these cases are especially prone to target--aspect binding failures and fabricated part-level targets.

\subsection{Collapsed Scalar Reward Variant}
\label{app:collapsed_reward}

To test whether explicit reward decomposition is necessary, we additionally consider a collapsed scalar reward variant that merges the main structural reward components into a single outcome-level score without component-specific weighting.
We use a simple uniform average so that the comparison isolates the value of explicit reward decomposition and component-specific weighting.

Specifically, we replace the decomposed reward in Eq.~\eqref{eq:reward_full} with
 \begin{equation}
\begin{aligned}
R_{\text{collapsed}}(\mathbf{x}, \hat{y})
&= \frac{1}{4} \Big(
R_{\text{base}}(\mathbf{x}, \hat{y})
+ R_{\text{bind}}(\mathbf{x}, \hat{y}) \\
&\quad
+ R_{\text{hall}}(\mathbf{x}, \hat{y})
+ R_{\text{cat}}(\mathbf{x}, \hat{y})
\Big).
\end{aligned}
\end{equation}

In this variant, we remove component-specific weighting and input-adaptive weighting, while keeping the rest of the on-policy optimization setup unchanged, including student rollouts, reverse-KL distillation, reference regularization, reward normalization, and the reward-to-weight transformation $w(\cdot)$.
The format-validity reward is omitted from this collapsed variant so that the comparison focuses on whether decomposing the main structural signals for binding, grounding, and category supervision provides better credit assignment than a single undifferentiated reward.

This variant preserves access to the same underlying structural information as \staropd{}, but discards the explicit decomposition used to bias updates toward specific failure modes.
As reported in Table~\ref{tab:ablation}, the weaker performance of this variant suggests that separating binding consistency, hallucination suppression, and category support yields more effective structural credit assignment than collapsing them into a single scalar outcome reward.

\subsection{Reward-to-Weight Transformation}
\label{app:reward_weight}
For each student rollout, STAR-OPD first computes the raw cascade-aware reward.
For brevity, $R_i = R(\mathbf{x}_i,\hat{y}_i)$ denotes the raw reward of the $i$-th rollout in the current batch.
Because the absolute reward scale may vary across batches, we normalize rewards within each batch before converting them into rollout weights.

Specifically, let $\mu_R$ and $\sigma_R$ denote the mean and standard deviation of raw rewards in the current batch.
We compute the normalized reward
 \begin{equation}
\tilde{R}_i = \frac{R_i - \mu_R}{\sigma_R + \epsilon},
\end{equation} 
where $\epsilon$ is a small constant for numerical stability.

The rollout weight is then defined as
 \begin{equation}
\alpha_i = w(R_i)
= \mathrm{clip}\!\left(
\exp(\tilde{R}_i / \tau_w),
\alpha_{\min},
\alpha_{\max}
\right),
\end{equation} 
where $\tau_w$ is a temperature parameter controlling the sharpness of reward weighting, and $\alpha_{\min}, \alpha_{\max}$ bound the rollout weight to avoid overly small or overly dominant updates.

Unless otherwise specified, we use $\tau_w = 2.0$, $\alpha_{\min}=0.2$, $\alpha_{\max}=3.0$, and $\epsilon = 10^{-6}$ in all experiments.
This transformation preserves the ordering of rollout quality, gives higher weight to better-structured student outputs, and improves training stability compared with using raw rewards directly.

\subsection{Training Dynamics of Reward Components}
\label{app:reward_dynamics}

\begin{figure}[t]
  \centering
  \includegraphics[width=0.98\columnwidth]{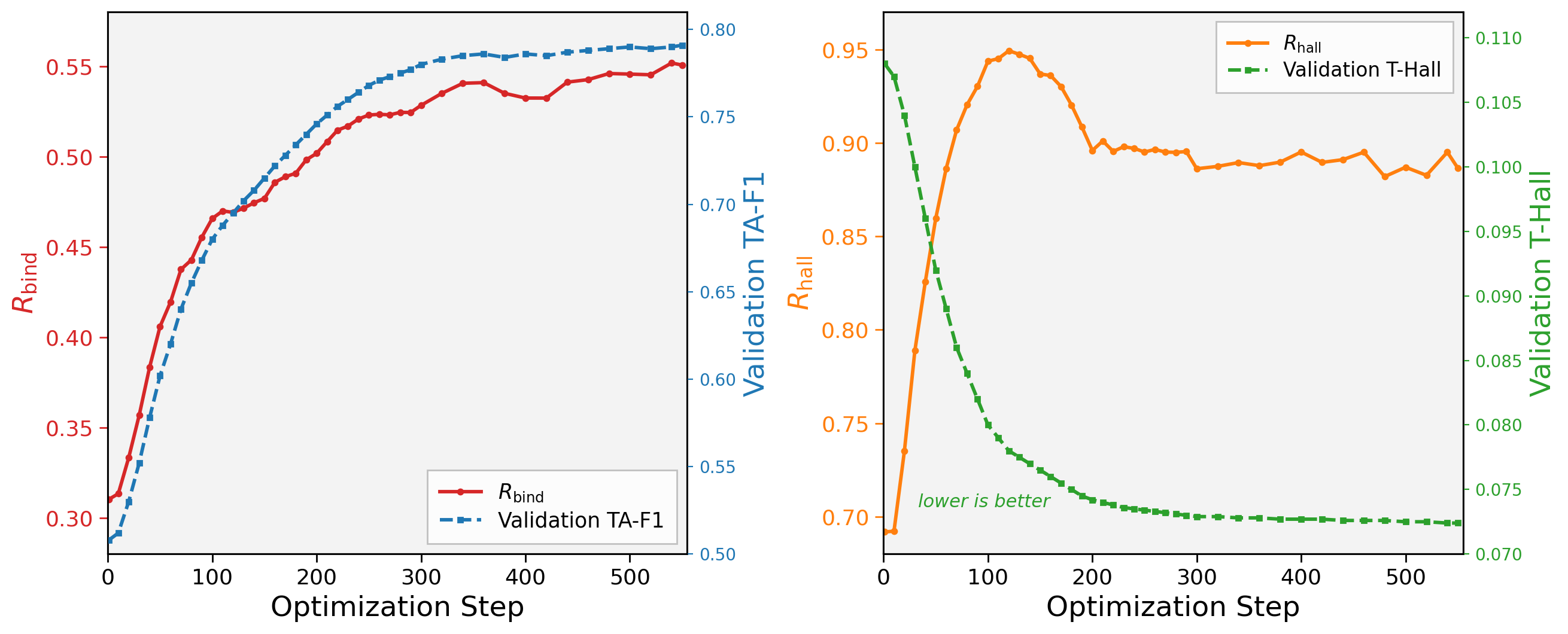}
  \caption{
    Training dynamics of the main reward components and corresponding validation structural metrics during \staropd{} optimization.
    Left: $R_{\text{bind}}$ and validation TA-F1.
    Right: $R_{\text{hall}}$ and validation T-Hall.
    The reward curves and validation metrics move in directions consistent with improved structural behavior and do not show obvious pathological late-stage divergence.
  }
  \label{fig:reward_dynamics}
\end{figure}

To examine whether the rule-based rewards remain aligned with the intended structural objectives during training, we track the dynamics of the two main reward components together with their corresponding validation metrics.
Figure~\ref{fig:reward_dynamics} plots $R_{\text{bind}}$ alongside validation TA-F1, and $R_{\text{hall}}$ alongside validation T-Hall.

The increase in $R_{\text{bind}}$ is accompanied by improved validation TA-F1, which is consistent with better target--aspect structural consistency.
Similarly, $R_{\text{hall}}$ rises sharply early in training and later fluctuates within a stable high-reward range, while validation T-Hall continues to decrease, indicating improved suppression of non-grounded target predictions.

We do not observe obvious late-stage reward inflation or instability together with metric degradation.
Although this analysis does not formally exclude shortcut optimization, it provides evidence that the reward terms remain broadly aligned with the structural behaviors they are intended to encourage.
This interpretation is also consistent with the strongest gains of \staropd{} on product-part and other structurally difficult subsets in Table~\ref{tab:hard}.

\subsection{Difficulty-Aware Sampling and Replay}

To improve hard-case coverage under on-policy training, we use a lightweight difficulty-aware sampling strategy.
Each training instance is assigned a difficulty score

\begin{equation}
d(x) = \mathbf{1}[\text{PP}] + \mathbf{1}[\text{Sarc}] + \mathbf{1}[\text{Cont}] + \mathbf{1}[\text{Impl}],
\end{equation} 

where each indicator is derived from teacher pseudo-labels and lightweight heuristic signals corresponding to the four hard phenomena; no diagnostic evaluation labels are used for training.
Sampling probability is then defined as

\begin{equation}
p(x) \propto 1 + \gamma \cdot d(x),
\end{equation} 
with $\gamma=0.5$ by default. 
Harder instances are sampled more frequently than easy ones.

We maintain a replay buffer of low-reward student rollouts to revisit unresolved structural failures.
A rollout $(x_i,\hat{y}_i)$ is added to the buffer if its normalized reward $\tilde{R}_i < -0.5$.
The buffer stores up to 4{,}096 rollouts in a FIFO manner.
During training, 20\% of each batch is drawn from the replay buffer when available, and the remaining 80\% is sampled from the training set.

These replayed rollouts typically correspond to unresolved structural failures, such as incorrect target--aspect bindings or hallucinated targets, and help focus updates on the student states where structure-aware feedback is most needed.

\subsection{Curriculum}

To stabilize optimization, we use a simple curriculum on hard-sample exposure.
During the first 30\% of training steps, the probability of sampling from the difficulty-aware distribution is increased linearly from 0.3 to 0.7; afterward, it is fixed at 0.7 for the remainder of training.

This curriculum reduces early optimization noise while ensuring that the model eventually receives sufficient coverage of structurally difficult cases.

\subsection{Optimization Hyperparameters}

Unless otherwise specified, all student methods use Qwen3-4B as the default student and Qwen3-1.7B for scale analysis.
On-policy distillation is run for 2K update iterations.

Unless otherwise specified, each input contributes one student rollout per update step.
We use a per-device training batch size of 4 with gradient accumulation over 2 steps on 4 GPUs, resulting in an effective batch size of 32 training instances per parameter update.
The maximum generation length for student rollouts is set to 1024 new tokens.

We use AdamW with a peak learning rate of \(1\times10^{-5}\), weight decay of 0.01, and a cosine decay schedule with 5\% warmup.
The reference-policy regularization coefficient in Eq.~\eqref{eq:loss} is set to \(\beta=0.1\).

For rollout generation, we use temperature 0.8 and top-\(p\) sampling with \(p=0.95\).
Unless otherwise noted, all experiments use seed 42.
For $R_{\text{cat}}$, teacher scoring is performed only at the aspect prediction step for the sampled student rollout, rather than over the full decoded sequence, to limit additional training overhead.

\subsection{Hardware and Runtime}

Teacher fine-tuning is conducted on 4$\times$A100 80G GPUs for approximately 14 hours.
On-policy student distillation is conducted on 4$\times$A100 80G GPUs for about 24 hours, with teacher inference served by vLLM \citep{kwon2023efficient}.
These numbers are reported to provide an approximate training-cost reference for \staropd{}; inference-time efficiency comparisons are given separately in Appendix~\ref{app:runtime}.

For evaluation, all reported numbers are averaged over three runs when applicable, and statistical significance is tested with paired bootstrap resampling at \(p<0.05\).

\subsection{Inference Efficiency}
\label{app:runtime}

To quantify the deployment advantage of distilled students, we measure inference efficiency on \eabsa{} Test under the same decoding configuration for teacher and student models.
Table~\ref{tab:runtime} reports the average end-to-end latency per review and review throughput on a single A100 80G GPU.

The Qwen3-4B student distilled with \staropd{} processes reviews 2.18$\times$ faster than the Qwen3-32B teacher, while substantially narrowing the performance gap.
The smaller 1.7B student is even faster, confirming the practical quality--efficiency advantage of distilled models for large-scale deployment.


\begin{table}[t]
  \centering
  \small
  \caption{
    Inference efficiency on \eabsa{} Test under the same decoding configuration.
    All models are evaluated on a single A100 80G GPU.
  }
  \label{tab:runtime}
  \setlength{\tabcolsep}{4pt}
  \begin{tabular}{lccc}
    \toprule
    \textbf{Model} & \textbf{Latency} & \textbf{Reviews/s} & \textbf{Speedup} \\
    \midrule
    Teacher$^{\dag}$            & 2.119 & 0.47 & 1.0$\times$ \\
    \staropd{} (4B)$^{\ddag}$   & 0.971 & 1.03 & 2.18$\times$ \\
    \staropd{} (1.7B)$^{\ddag}$ & 0.290 & 3.45 & 7.31$\times$ \\
    \bottomrule
  \end{tabular}
  \vspace{2pt}
  \caption*{\scriptsize
  $^{\dag}$ Qwen3-32B teacher model.\quad
  $^{\ddag}$ Qwen3 student models distilled via \staropd{}.}
\end{table}

\section{Training Algorithm}
\label{app:algo}
\begin{algorithm}[h]
\small
\caption{\textbf{\staropd{}}: On-Policy Cascade-Aware Distillation}
\label{alg:quadopd}
\begin{algorithmic}[1]
\Require Teacher policy $\pi_T$, initial student policy $\pi_S$, reference policy $\pi_{\mathrm{ref}} \leftarrow \pi_S$, training set $\mathcal{D}$
\State Initialize replay buffer $\mathcal{B} \leftarrow \emptyset$
\For{each training iteration}
    \State Sample a batch $\mathcal{S}$ from $\mathcal{D}$ with difficulty-aware sampling
    \State Optionally mix in low-reward samples from $\mathcal{B}$
    \For{each input $\mathbf{x}_i \in \mathcal{S}$}
        \State Sample a student rollout $\hat{y}_i \sim \pi_S(\cdot \mid \mathbf{x}_i)$
        \State Compute cascade-aware reward $R_i$ using $R_{\text{base}}, R_{\text{bind}}, R_{\text{hall}}, R_{\text{cat}}, R_{\text{fmt}}$
        \If{$R_i$ is low}
            \State Add $(\mathbf{x}_i, \hat{y}_i)$ to $\mathcal{B}$
        \EndIf
    \EndFor
    \State Normalize rewards within the batch
    \State Convert normalized rewards to rollout weights $\alpha_i = w(R_i)$
    \State Compute the reward-weighted on-policy objective
    \[
      \mathcal{L} = \mathcal{L}_{\mathrm{rKL}} + \beta \mathcal{L}_{\mathrm{ref}}
    \]
    \State Update student parameters
\EndFor
\State \Return $\pi_S$
\end{algorithmic}
\end{algorithm}

\section{Additional Qualitative Cases}
\label{app:qual_cases_app}
We provide additional qualitative examples to complement the hard-case results in the main text.
The cases below cover all four hard phenomena used in \eabsa{}-Hard: product-part mentions, colloquial sarcasm, contrastive sentences, and implicit opinion expressions.
Targets follow the same normalized schema used throughout the paper:
\texttt{PRODUCT} denotes an implicit product-level target,
\texttt{PRODUCT:xx} denotes an explicitly mentioned product target,
and \texttt{\pp{}:xx} denotes an explicitly mentioned product-part.

Across these cases, SeqKD typically remains locally plausible but drifts in target--aspect structure once generation departs from teacher states.
MiniLLM is more tolerant to local deviations, yet often backs off to coarser target/aspect structures or produces plausible but non-grounded targets.
In contrast, \staropd{} more reliably preserves target grounding, fine-grained aspect bindings, and complete tuple structure.

\begin{figure*}[t]
\centering
\scriptsize
\setlength{\tabcolsep}{4pt}
\renewcommand{\arraystretch}{1.03}

\begin{tabularx}{0.98\textwidth}{>{\raggedright\arraybackslash}p{0.11\textwidth} X}
\toprule
\rowcolor{gray!10}
\multicolumn{2}{c}{\textbf{(a) product-part (womenbags)}} \\
\midrule
\textbf{Review} &
The bag looks elegant, but the zipper got stuck on the second day. \\
\textbf{Gold} &
(\texttt{PRODUCT:bag}, \texttt{Design Aesthetic}, \textit{looks elegant}, \texttt{pos});
(\texttt{\pp{}:zipper}, \texttt{Hardware Quality}, \textit{got stuck}, \texttt{neg}) \\
\textbf{SeqKD} &
(\texttt{\pp{}:zipper}, \textcolor{darkred}{\texttt{Ease of Use}}, \textit{got stuck}, \texttt{neg})
{\footnotesize\textcolor{darkred}{ [binding drift]}} \\
\textbf{MiniLLM} &
(\textcolor{darkred}{\texttt{\pp{}:handle}}, \texttt{Hardware Quality}, \textit{got stuck}, \texttt{neg})
{\footnotesize\textcolor{darkred}{ [hallucinated target]}} \\
\textbf{\staropd{}} &
(\texttt{\pp{}:zipper}, \texttt{Hardware Quality}, \textit{got stuck}, \texttt{neg})
{\footnotesize\textcolor{darkgreen}{ [correct grounding + binding]}} \\
\bottomrule
\end{tabularx}

\vspace{0.35em}

\begin{tabularx}{0.98\textwidth}{>{\raggedright\arraybackslash}p{0.11\textwidth} X}
\toprule
\rowcolor{gray!10}
\multicolumn{2}{c}{\textbf{(b) Contrast (furniture)}} \\
\midrule
\textbf{Review} &
The chair looks modern and fits my desk perfectly, but the seat feels hard and the wheels keep squeaking. \\
\textbf{Gold} &
(\texttt{PRODUCT:chair}, \texttt{Appearance \& Design}, \textit{looks modern}, \texttt{pos});
(\texttt{PRODUCT:chair}, \texttt{Fit \& Space Compatibility}, \textit{fits my desk perfectly}, \texttt{pos}); \newline
(\texttt{\pp{}:seat}, \texttt{Comfort}, \textit{feels hard}, \texttt{neg});
(\texttt{\pp{}:wheels}, \texttt{Noise Performance}, \textit{keep squeaking}, \texttt{neg}) \\
\textbf{SeqKD} &
(\texttt{\pp{}:seat}, \textcolor{darkred}{\texttt{Ergonomics}}, \textit{feels hard}, \texttt{neg});
(\texttt{\pp{}:wheels}, \textcolor{darkred}{\texttt{Ergonomics}}, \textit{keep squeaking}, \texttt{neg})
{\footnotesize\textcolor{darkred}{ [error propagation]}} \\
\textbf{MiniLLM} &
(\texttt{PRODUCT:chair}, \texttt{Comfort}, \textit{feels hard}, \texttt{neg});
(\texttt{PRODUCT:chair}, \texttt{Noise Performance}, \textit{keep squeaking}, \texttt{neg})
{\footnotesize\textcolor{darkred}{ [target collapse]}} \\
\textbf{\staropd{}} &
(\texttt{\pp{}:seat}, \texttt{Comfort}, \textit{feels hard}, \texttt{neg});
(\texttt{\pp{}:wheels}, \texttt{Noise Performance}, \textit{keep squeaking}, \texttt{neg})
{\footnotesize\textcolor{darkgreen}{ [full structure preserved]}} \\
\bottomrule
\end{tabularx}

\caption{Additional qualitative cases: product-part and contrastive reviews.}
\label{fig:qual_cases_app1}
\end{figure*}

\begin{figure*}[t]
\centering
\scriptsize
\setlength{\tabcolsep}{4pt}
\renewcommand{\arraystretch}{1.03}

\begin{tabularx}{0.98\textwidth}{>{\raggedright\arraybackslash}p{0.11\textwidth} X}
\toprule
\rowcolor{gray!10}
\multicolumn{2}{c}{\textbf{(c) Sarcasm (makeup)}} \\
\midrule
\textbf{Review} &
Love how this ``long-lasting'' lipstick vanishes before I finish my coffee. \\
\textbf{Gold} &
(\texttt{PRODUCT:lipstick}, \texttt{Longevity}, \textit{vanishes before I finish my coffee}, \texttt{neg}) \\
\textbf{SeqKD} &
(\texttt{PRODUCT:lipstick}, \textcolor{darkred}{\texttt{Longevity}}, \textcolor{darkred}{\textit{long-lasting}}, \textcolor{darkred}{\texttt{pos}})
{\footnotesize\textcolor{darkred}{ [surface cue imitation]}} \\
\textbf{MiniLLM} &
(\texttt{PRODUCT:lipstick}, \textcolor{darkred}{\texttt{Overall}}, \textit{vanishes before I finish my coffee}, \texttt{neg})
{\footnotesize\textcolor{darkred}{ [coarse aspect fallback]}} \\
\textbf{\staropd{}} &
(\texttt{PRODUCT:lipstick}, \texttt{Longevity}, \textit{vanishes before I finish my coffee}, \texttt{neg})
{\footnotesize\textcolor{darkgreen}{ [fine-grained recovery]}} \\
\bottomrule
\end{tabularx}

\vspace{0.35em}

\begin{tabularx}{0.98\textwidth}{>{\raggedright\arraybackslash}p{0.11\textwidth} X}
\toprule
\rowcolor{gray!10}
\multicolumn{2}{c}{\textbf{(d) Implicit opinion (dresses)}} \\
\midrule
\textbf{Review} &
It zipped up fine, but I couldn't sit down in it. \\
\textbf{Gold} &
(\texttt{PRODUCT}, \texttt{Ease of Use}, \textit{zipped up fine}, \texttt{pos});
(\texttt{PRODUCT}, \texttt{Comfort}, \textit{couldn't sit down in it}, \texttt{neg}) \\
\textbf{SeqKD} &
(\texttt{PRODUCT}, \textcolor{darkred}{\texttt{Fit}}, \textit{couldn't sit down in it}, \texttt{neg})
{\footnotesize\textcolor{darkred}{ [neighboring aspect confusion]}} \\
\textbf{MiniLLM} &
(\texttt{PRODUCT}, \textcolor{darkred}{\texttt{Overall Satisfaction}}, \textit{couldn't sit down in it}, \texttt{neg})
{\footnotesize\textcolor{darkred}{ [coarse aspect fallback]}} \\
\textbf{\staropd{}} &
(\texttt{PRODUCT}, \texttt{Comfort}, \textit{couldn't sit down in it}, \texttt{neg})
{\footnotesize\textcolor{darkgreen}{ [correct implicit mapping]}} \\
\bottomrule
\end{tabularx}

\caption{Additional qualitative cases: sarcasm and implicit opinion reviews.}
\label{fig:qual_cases_app2}
\end{figure*}

These examples mirror the quantitative hard-case trends in the main text.
Product-part cases and contrastive sentences highlight structural target--aspect failures, including binding drift, target collapse, and error propagation across tuples.
The sarcasm and implicit  opinion expressions further show that the gains of \staropd{} do not come only from better local sentiment prediction, but also from more stable recovery of fine-grained aspect structure under ambiguous evidence.

\section{Full Prompt Template}
\label{app:prompt}
{\small
\begin{verbatim}
You are an expert in fine-grained sentiment
analysis for e-commerce product reviews.

Task: Extract all aspect sentiment quadruples.
Format: (target, aspect_category, opinion, sentiment)

Target hierarchy:
  PRODUCT          - product in general
  PRODUCT:XX       - specific product type
  PRODUCT_PART:XX  - specific product part
  !! Do NOT invent targets absent from text.

Aspect: one from the predefined list.
Opinion: minimal exact phrase from the review.
Sentiment: positive | neutral | negative

Rules:
1. Extract every applicable (target,aspect) pair.
2. One quadruple per (target,aspect) combination.
3. Keep opinion as shortest faithful text span.
4. Handle negation: do not assign positive
   sentiment to negated phrases.
5. Handle concession: assign separate quadruples
   with correct polarity to each clause.
6. Multi-product: assign each opinion to
   the correct target product.
7. Verify: all targets exist in review text;
   all aspects from predefined list only.

<thinking>
Step 1: List all target entities in the review.
Step 2: Map each target to aspect categories.
Step 3: Locate minimal opinion span for each.
Step 4: Resolve sentiment under negation/concession.
Step 5: Verify no invented targets;
        all aspects in predefined list.
</thinking>

<quadruples>
[one quadruple per line]
</quadruples>

Review: "{{review_text}}"
Aspect list: {{category_list}}
\end{verbatim}
}

\section{Aspect Taxonomy Example}
\label{app:taxonomy}

We provide the women's bags taxonomy as a representative example of the fine-grained aspect space used in \eabsa{}.
The full flat list is provided to the model without grouping information.

\begin{table}[h]
  \centering
  \small
  \caption{Aspect taxonomy for women's bags (34 categories).}
  \label{tab:taxonomy}
  \setlength{\tabcolsep}{3pt}
  \begin{tabular}{ll}
    \toprule
    \textbf{Theme} & \textbf{Aspect Categories} \\
    \midrule
    Overall    & Overall Satisfaction \\
    Material   & Material Type, Material Quality, \\
               & Material Accuracy, Water Resistance, \\
               & Odor, Cleaning \& Maintenance \\
    Appearance & Design Aesthetic, Color Accuracy, \\
               & Color Preference, Appearance \\
    Size       & Size Accuracy, Weight, \\
               & Capacity, Shape Retention \\
    Hardware   & Hardware Quality, Hardware Design, \\
               & Security Features \\
    Strap      & Strap Adjustability, Comfort \\
    Interior   & Interior Organization \\
    Build      & Construction Quality, Durability \\
    Commerce   & Price, Value for Money, \\
               & Customer Service, Shipping Speed, \\
               & Delivery Experience, Packaging, \\
               & Return \& Refund Policy \\
    Context    & Occasion Suitability, \\
               & Brand Authenticity, \\
               & Sizing Guidance Clarity, \\
               & Accessories Included \\
    \bottomrule
  \end{tabular}
\end{table}


\end{document}